\documentclass[10pt,twocolumn,letterpaper]{article}

\usepackage{cvpr} 
\usepackage{times}
\usepackage{epsfig}
\usepackage{graphicx}
\usepackage{amsmath}
\usepackage{amssymb}

\usepackage{multirow}
\usepackage{graphicx}
\usepackage{placeins}
\usepackage{float} 
\usepackage[table]{xcolor}  
\usepackage{colortbl}
\usepackage{dblfloatfix}

\begin{document}

\title{NLML-HPE: Head Pose Estimation with Limited Data via Manifold Learning}

\author{Mahdi Ghafourian, Federico M. Sukno\\
Department of Engineering, Universitat Pompeu Fabra, Spain\\
{\tt\small (mahdi.ghafourian,federico.sukno)@upf.edu}\\
}

\maketitle
\thispagestyle{empty}

\begin{abstract}
   Head pose estimation (HPE) plays a critical role in various computer vision applications such as human-computer interaction and facial recognition. In this paper, we propose a novel deep learning approach for head pose estimation with limited training data via non-linear manifold learning called NLML-HPE. This method is based on the combination of tensor decomposition (i.e., Tucker decomposition) and feed forward neural networks. Unlike traditional classification-based approaches, our method formulates head pose estimation as a regression problem, mapping input landmarks into a continuous representation of pose angles. To this end, our method uses tensor decomposition to split each Euler angle (yaw, pitch, roll) to separate subspaces and models each dimension of the underlying manifold as a cosine curve. We address two key challenges: 1. Almost all HPE datasets suffer from incorrect and inaccurate pose annotations. Hence, we generated a precise and consistent 2D head pose dataset for our training set by rotating 3D head models for a fixed set of poses and rendering the corresponding 2D images. 2. We achieved real-time performance with limited training data as our method accurately captures the nature of rotation of an object from facial landmarks. Once the underlying manifold for rotation around each axis is learned, the model is very fast in predicting unseen data. Our training and testing code is available online along with our trained models: \url{https://github.com/MahdiGhafoorian/NLML_HPE}.
   
\end{abstract}

\section{Introduction}
Head pose estimation (HPE) refers to the process of determining the orientation and in some cases, the position of a person’s head relative to a camera or a global coordinate system~\cite{abate2022head, khan2021head}. Accurate HPE is crucial for various applications, including driver assistance (monitoring attention), human-computer interaction (enhancing interfaces), virtual reality (improving perspective rendering), surveillance (behavior analysis), and targeted advertising (tracking visual attention)~\cite{asperti2023deep}.

\begin{figure}[htbp]
 \centering 
 \includegraphics[trim={7cm 0cm 0 0cm},clip,width=120mm,scale=0.5]{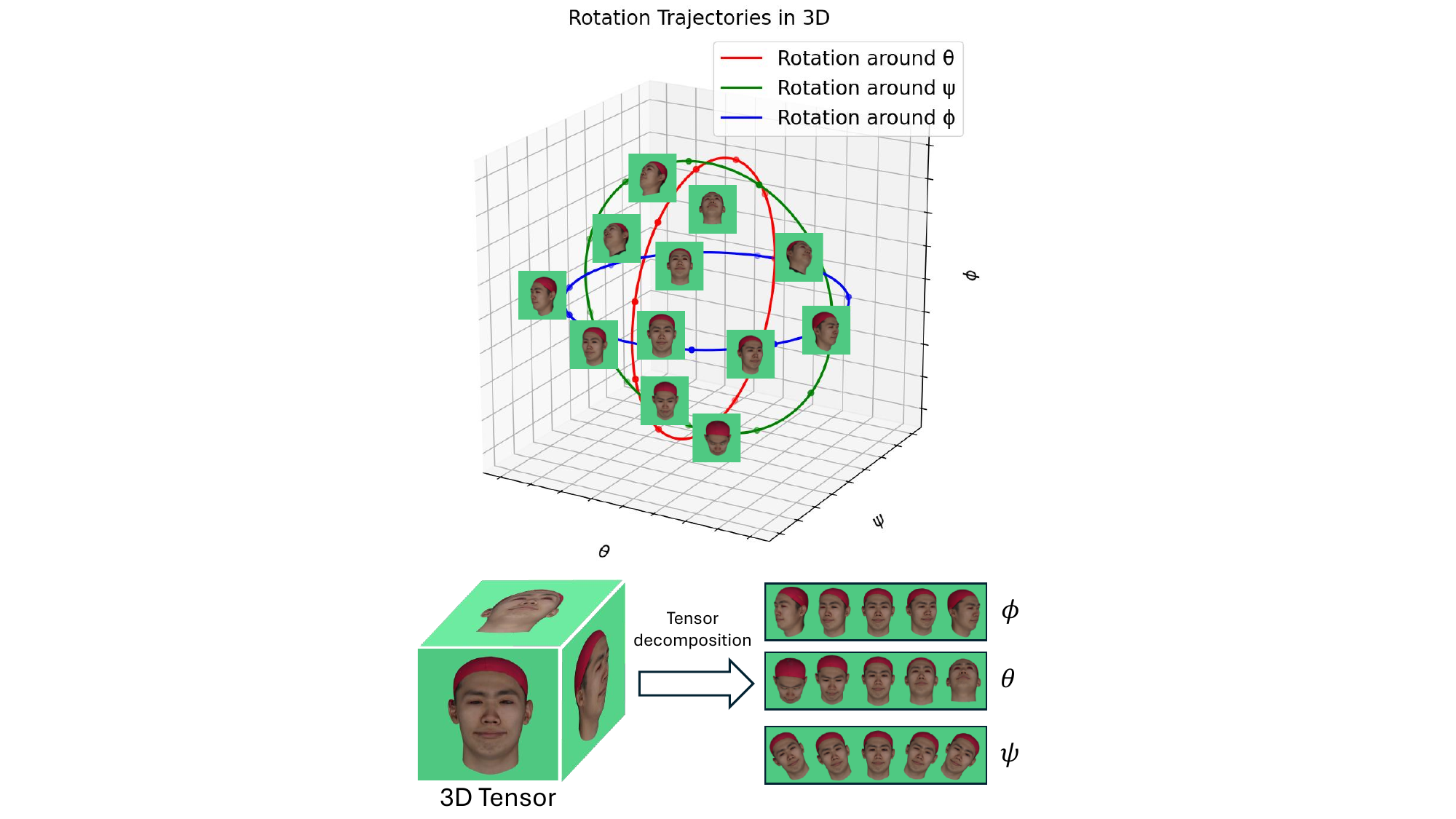}
 \caption{Overview of the decomposition of head pose variation from a sample of our generated dataset into separate subspaces.}
 \label{fig:Intro}
\end{figure}

There are various ways to represent HPE. One of the most popular representations is in terms of Euler angles. This representation is commonly expressed with three angles: \textit{yaw}, or rotating around the vertical axis (e.g., left or right head movement), \textit{pitch}, or rotating around the lateral axis (e.g., up or down head movement), and \textit{roll}, or rotating around the longitudinal axis (e.g., tilting of the head sideways). Although a face image is considered as a high-dimensional data sample, a key observation is that various poses of that can define a low-dimensional manifold which is constrained by the allowable pose variations. As a result, head pose estimation can be viewed as a problem of dimensionality reduction and manifold learning~\cite{takallou2014head, derkach2019tensor}.

Dimension reduction techniques can generally be categorized into linear, multi-linear, and non-linear methods. Among the most widely used linear approaches are Principal Component Analysis (PCA)~\cite{turk1991face} and Linear Discriminant Analysis (LDA)~\cite{lu2003face}. Early studies~\cite{yang2004two, wan2011feature} have leveraged PCA and LDA for face recognition tasks~\cite{ghafourian2023otb, ghafourian2023toward}. Another notable linear method, Locality Preserving Projections (LPP)~\cite{chen2008face}, along with its variants~\cite{lu2011face, soundar2010preserving}, focuses on maintaining local relationships within the facial dataset while capturing its intrinsic manifold structure. However, due to the non-linearity of the pose manifold, linear methods struggle to accurately reveal the underlying manifold.

Multi-linear analysis, on the other hand, extends linear methods by providing a mathematical framework well-suited for dimensionality reduction in multi-mode data problems. In such cases, data are represented as multi-dimensional tensors, and multi-linear decomposition is employed to break down the tensor into its orthogonal mode matrices. Methods like Isomap~\cite{raytchev2004head} and Locally Linear Embedding (LLE)~\cite{fu2006graph} have been investigated to learn the underlying manifold structure defined by orientation parameters. Although these techniques can learn low-dimensional representations of data, the manifold they produce is defined implicitly. This makes it difficult to impose specific constraints that accurately model the inherent structure induced by rotation variations.


In this paper, we propose a non-linear, tensor-based method grounded in multi-linear decomposition~\cite{de2000multilinear, wang2017learning} to learn the manifold defined by 3D rotation. In particular, our method is able to map the underlying structure of high-dimensional face images into low-dimensional pose manifolds, enabling learning rotation around each Euler angles as a separated manifold. To do this, we use multi-linear decomposition (i.e., Tucker decomposition) over data descriptors (i.e., facial landmarks) to split rotation factors (i.e., yaw, pitch, roll) and obtain a set of subspaces. Fig.~\ref{fig:Intro} shows a visualization of such a decomposition for a $\mathbb{R}^{3}$ tensor. The principal components of these subspaces correspond to the pose variation around one of the Euler angles and defines a continuous curves that are governed by unique sinusoidal parameters. Tensor decomposition is a computationally expensive operation, requiring significant processing time and memory, which makes it unsuitable for real-time inference. To overcome this limitation, we incorporate an encoder with three Multi-Layer Perceptron (MLP) heads, each responsible for predicting one Euler angle. The advantage of our approach over using tensor decomposition alone is that we perform the decomposition on a limited subset of the input data and then train a lightweight encoder and MLP heads to achieve equivalent inference results in real time.

To perform tensor decomposition, input data in the training set must be pose consistent. This means that for proper decomposition of the tensor, there must be exactly one input sample for each possible combination of yaw, pitch, and roll to populate the tensor. To the best of the authors’ knowledge, no existing HPE dataset possesses this characteristic. Although it is possible to populate the tensor via interpolation for the missing combinations, the distribution of pose configurations in the training set may result in a tensor that is predominantly filled with interpolated features, leading to inaccurate eigenvectors. Consequently, instead of capturing the principal features of the data, the eigenvectors might primarily reflect random fluctuations introduced by the interpolation process. In our experiments, we address this issue by generating a pose-consistent dataset through the intrinsic rotation of 3D models from the FaceScape dataset~\cite{yang2020facescape} for each pose combination and rendering the corresponding 2D images. This approach not only resolves the pose-consistency problem but also produces a highly accurate annotated HPE dataset, in contrast to the error-prone manual annotations found in other datasets. We perform our pose estimation experiments using our generated dataset for decomposition and for training the encoder. We employ the MediaPipe Face Mesh Toolkit~\cite{lugaresi2019mediapipe} to extract facial landmarks from our generated face images. We evaluate our approach on two popular, publicly available HPE datasets, BIWI~\cite{fanelli2013random} and AFLW2000~\cite{zhu2016face}, and demonstrate that angle estimation can be achieved by training three MLP heads rather than through the time-consuming optimization of cosine functions as described in~\cite{derkach2019tensor}. This enables both state-of-the-art accuracy and real-time performance.

The rest of this paper is organized as follows. Section~\ref{sec:related_work} provides a brief review of the existing head pose estimation approaches. Section~\ref{sec:background} presents an overview of tensor decomposition methods, with a particular emphasis on higher-order singular value decomposition (HOSVD). In Section~\ref{sec:method}, we introduce our NLML-HPE framework, which models data variations caused by rotations for real-time pose estimation. Section~\ref{sec:experiments} presents experiments conducted on two widely used HPE datasets, along with our generated dataset. Section~\ref{sec:limitations} discusses the potential limitations of our approach. Finally, Section~\ref{sec:conclusion} concludes the paper.

\section{Related Work}
\label{sec:related_work}
Human pose estimation is a widely studied task in computer vision with very diverse approaches throughout its history. Therefore, we only focus on those approaches that are related to manifold learning as an extensive review of all HPE method is beyond of the scope of this paper. 

The main idea behind manifold-based methods is to consider modeling the underlying structure of head pose variations~\cite{wang2017head}. In fact, the set of all facial images generated by varying the orientation of a face is intrinsically a three-dimensional manifold (ignoring or compensating for other types of image
variation like changes in scale, illumination etc.), which however is embedded in image space of a much higher dimensionality~\cite{raytchev2004head}. Due to the inherent non-linearity of this manifold within the feature-defined ambient space, researchers have investigated non-linear manifold learning methods such as Locally Linear Embedding~\cite{fu2006graph}, Isomap~\cite{raytchev2004head}, Synchronized Submanifold Embedding~\cite{zhu2014automatic}, and Homeomorphic Manifold Analysis~\cite{peng2014head}. In 2014, Takallou and Kasaei~\cite{takallou2014head} proposed a non-linear tensor model for head pose estimation, using multi-linear decomposition to capture identity, pose, and pixel information, with a focus on yaw rotations. Their method projects query images into pose and identity subspaces, generating pose parameters, which are validated against a unified pose manifold to obtain the final estimate. Sundararajan and Woodard~\cite{sundararajan2015head} introduced a method for head pose estimation in unconstrained images using feature-based manifold embedding. They addressed the challenge of learning a similarity kernel that reflects variations solely due to head pose, while disregarding other sources of variation. To achieve this, they utilized identity-invariant Geometric Blur features to learn the similarity kernel. Later in 2018, Hong et al.~\cite{hong2018multimodal} introduced a novel framework for face pose estimation that integrates multi-modal data and multitask learning. They enhanced traditional convolutional neural networks by incorporating Manifold Regularized Convolutional Layers (MRCL), which learn the relationships among neuron outputs in a low-rank space. 

The most similar work to ours is that of Derkach et al.~\cite{derkach2019tensor}, who utilized multi-linear decomposition to separate pose variations (yaw, pitch, and roll) into distinct subspaces, effectively modeling the underlying 3D manifold resulting from these rotations. Their method comprises two phases: (i) a training phase where they conduct HOSVD to decompose the populated tensor and optimize the trigonometric parameters; and (ii) a test phase where they optimize variation angles for the input face image by computing \( \underset{w^{(\text{yaw})}, w^{(\text{pitch})}, w^{(\text{roll})}}{\arg\min} \left\lVert x - \hat{x} \right\rVert \) where $\hat{x}$ is obtained via Einstein summation~\cite{albert1916foundation} operation that is computationally expensive. This method is not feasible for real-time applications. In contrast, we present a method that employs a lightweight encoder with three MLP heads to obtain head pose angles for the input image instantaneously.

\section{Technical Background: Multi-linear Decomposition}
\label{sec:background}
Multi-linear Decomposition refers to the process of decomposing multi-dimensional data into simpler, lower-dimensional components. One common method for multi-linear decomposition is tensor decomposition, such as Tucker decomposition or CANDECOMP/PARAFAC (CP). Higher order SVD (HOSVD)  is a type of Tucker decomposition where the factor matrices are found by applying Singular Value Decomposition (SVD) to each unfolded slice (mode) of the tensor~\cite{bergqvist2010higher, comon2014tensors}. In particular, a tensor is often referred to as an n-way array or n-mode matrix. For example, vectors and tensors can be viewed as first- and second-order tensors, respectively. To enhance understanding, we first revisit the standard SVD.

Given matrix $A \in \mathbb{R}^{m\times n}$, the SVD is as follows: 
\begin{equation}
A = U{\Sigma}{V^T} = \sum_{k=1}^{r} \sigma_k u_k v_k^T = \sum_{k=1}^{r} \sigma_k u_k \otimes v_k
\label{eq:svd_approx1}
\end{equation}
That is for the element $A_{ij}$ of $A$,
\begin{equation}
a_{ij} = \sum_{k=1}^{r} U_{ik}\Sigma_{kk}V_{jk} = \sum_{k=1}^{r} \sigma_k U_k V_{jk}
\label{eq:svd_approx2}
\end{equation}
where $\otimes$ denotes the tensor (or outer) product $\mathbf{x} \otimes \mathbf{y} \triangleq \mathbf{x} \mathbf{y}^T; \Sigma$ is a diagonal $(r\times r)$ matrix with nonzero singular
values of $A$ (the square roots of the eigenvalues of $A^TA$) on
its diagonal; $u_k$ and $v_k$ are the orthonormal columns of the
matrix $U$ $(m\times r)$ and $V$ $(n \times r)$, respectively, with $v_k$ being
the eigenvectors of $A^TA$ and $u_k = Av_k/{\sigma}k$~\cite{bergqvist2010higher}.

The SVD is helpful when dealing with a two-dimensional dataset \( A_{ij} \), which is naturally represented as a matrix \( A \). In many applications, such as ours, we have to deal with multidimensional data. Particularly, we aim to model data that depends on five factors: features, identity, and three rotations (yaw, pitch, roll) about orthogonal spatial axes using a fifth-order tensor. This approach allows us to effectively capture the complex inter-dependencies among these factors. 

\begin{figure}[htbp]
 \centering 
 \includegraphics[trim={8cm 0.5cm 0 2cm},clip,width=125mm,scale=0.5]{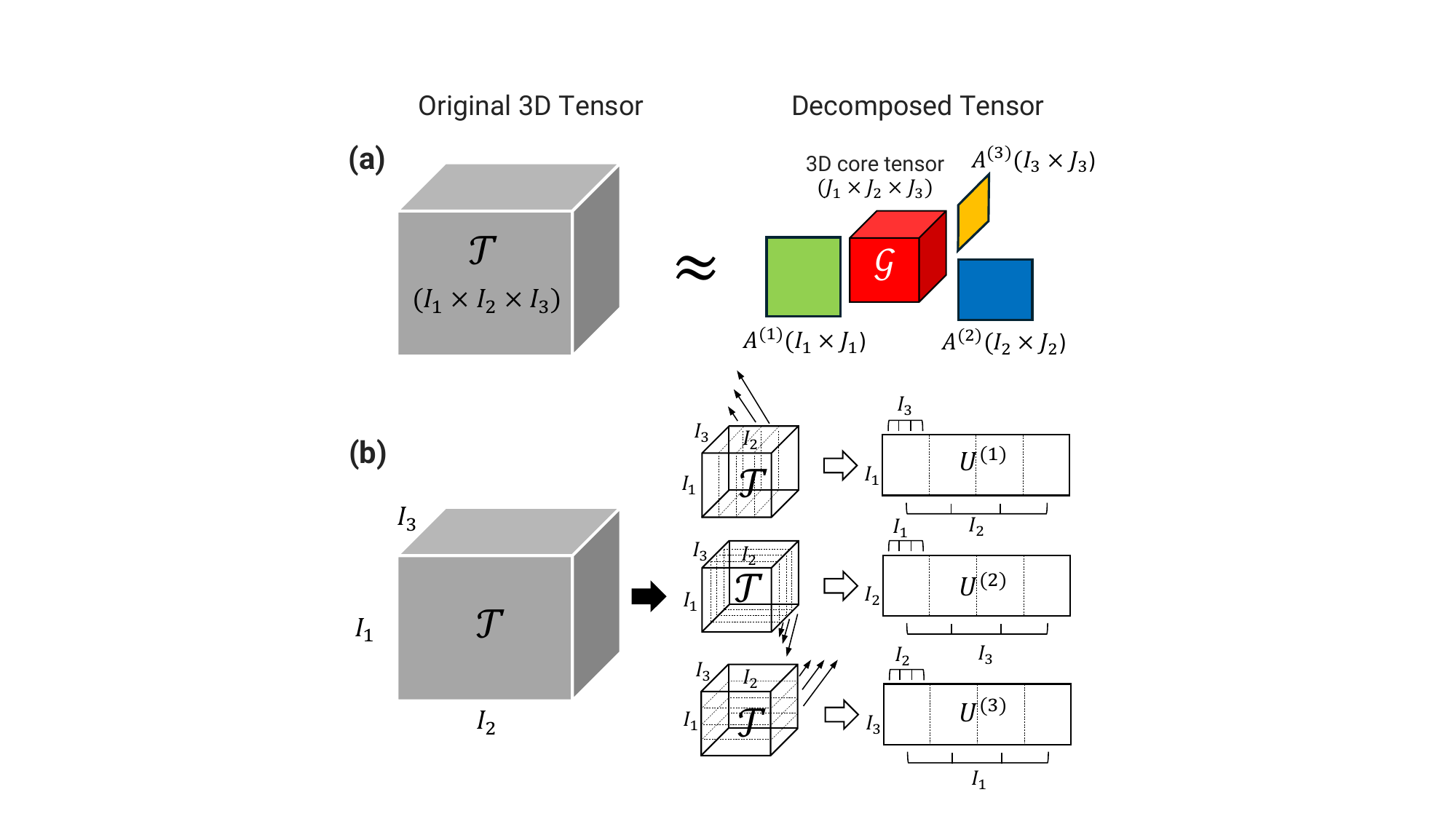}
 \caption{(a) depiction of a 3D tensor decomposition. (b) Unfolding of the $(I_1 \times I_2 \times I_3)$-tensor $T$ to the $(I_1 \times I_2 I_3)$-matrix $\mathbf{U}^{(1)}$, the $(I_2 \times I_3 I_1)$-matrix $\mathbf{U}^{(2)}$, and the $(I_3 \times I_1 I_2)$-matrix $\mathbf{U}^{(3)}$ $(I_1 = I_2 = I_3 = 4)$.}
 \label{fig:Tensor_decomposition}
\end{figure}

The SVD may be generalized to higher order tensors or multiway arrays in several ways. The approaches we use in this paper is so-called Tucker/HOSVD decomposition shown in Fig.~\ref{fig:Tensor_decomposition} (a).

Given a fifth order tensor $T\in \mathbb{R}^{I_1\times I_2\times I_3\times I_4\times I_5}$, the tucker decomposition can be expressed as:

\begin{equation}
T = \sum_{J_1=1}^{I_1} ... \sum_{J_5=1}^{I_5} G_{J_1J_2...J_5} a_{J_1}^{(1)}\otimes a_{J_2}^{(2)}\otimes...\otimes a_{J_5}^{(5)}
\label{eq:tucker_component}
\end{equation}
or with mode product~\cite{de2000multilinear} as:
\begin{equation}
T_{j_1j_2...j_5} = \sum_{J_1=1}^{I_1} ... \sum_{J_5=1}^{I_5} G_{J_1J_2...J_5}\times_1 A_{j_1J_1}^{(1)}\times_2...\times_5 A_{j_5J_5}^{(5)}
\label{eq:tucker_modeproduct}
\end{equation}
where $G\in \mathbb{R}^{J_1\times J_2\times J_3\times J_4\times J_5}$ is the core tensor that captures interactions between the components and \( A^{(n)} \in \mathbb{R}^{I_n \times J_n} \) are the factor matrices (i.e., principal components) associated with each mode \( n \) (for \( n = 1, 2, 3, 4, 5 \)), representing how the original tensor dimensions relate to the lower-dimensional core tensor. The n-mode product ($\times_{n}$) is an operation in tensor algebra that multiplies a tensor by a matrix along a specified mode (dimension). It generalizes matrix-vector and matrix-matrix multiplication to higher-order tensors. Therefore, n-mode product of $G$ with $A$ denoted as $G\times_{n}A$ results in a new tensor $W\in \mathbb{R}^{J_1\times J_2\times ...\times J_{n-1}\times I_n\times J_{n+1}\times ...\times J_N}$ and is defined element-wised as:
\begin{equation}
(G\times_{n}A)_{J_1...J_{n-1}I_{n}J_{n+1}...J_N}=\sum_{j_n=1}^{J_n}G_{j_{1}j_{2}...j_{N}}A_{i_{n}j_n}
\label{eq:tucker_unfolding}
\end{equation}
HOSVD is equivalent to applying SVD to each factor matrix $A^{n}$ obtained by unfolding (a.k.a matricization or flatenning) of $T$~\cite{bergqvist2010higher}. An example of unfolding a third-order tensor is illustrated in Fig.~\ref{fig:Tensor_decomposition} (b).

\section{Proposed Method}
\label{sec:method}

\subsection{Generation of pose-consistent dataset}
\label{subsec:pose-consistent}
\begin{figure*}[htp]
 \centering 
 \includegraphics[trim={0.8cm 2.5cm 0 1.5cm},clip,width=180mm,scale=0.5]{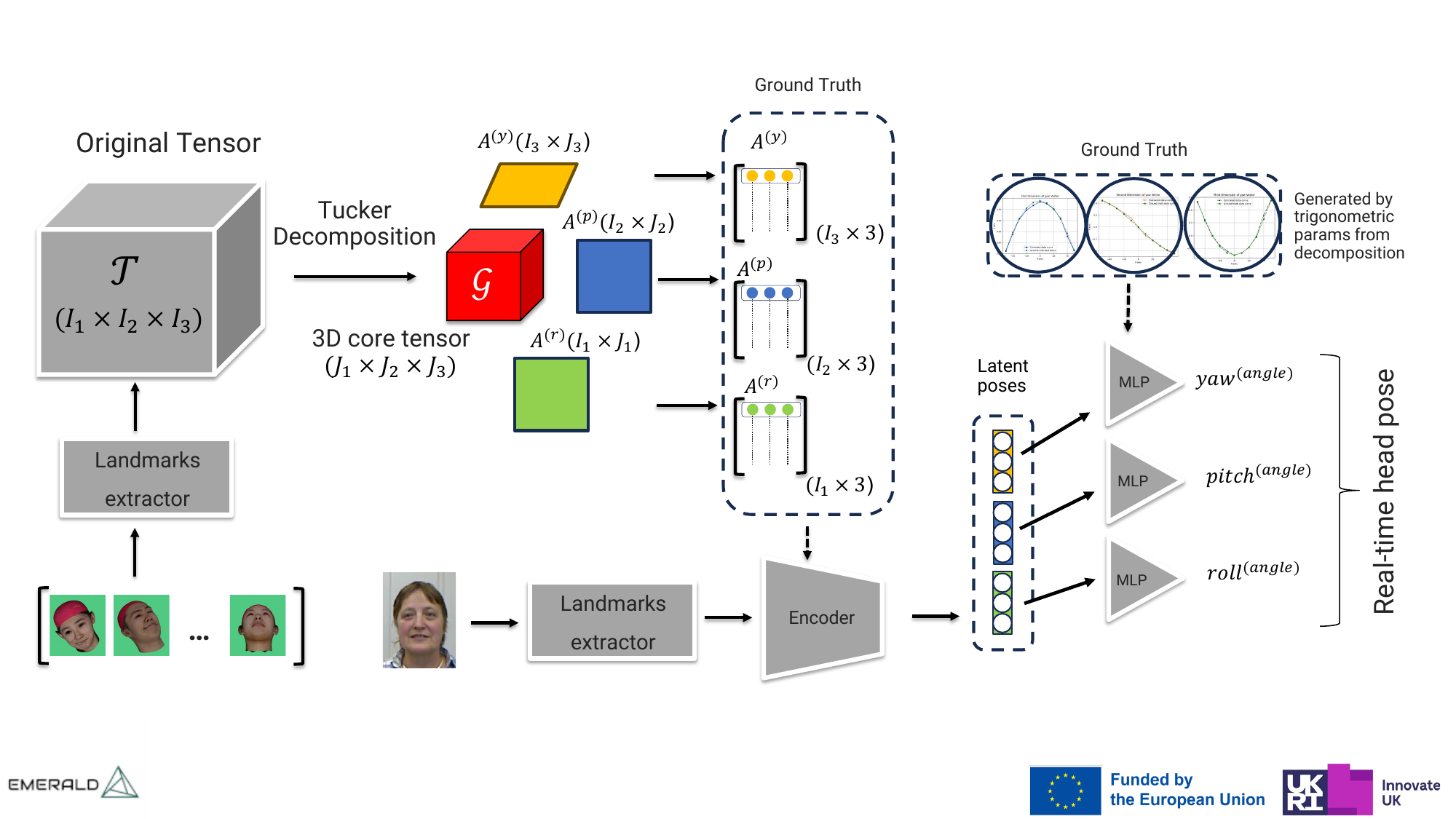}
 \caption{Overview of the architecture used in our proposed non-linear manifold learning framework.}
 \label{fig:Architecture}
\end{figure*}

To train the NLML-HPE method shown in Fig.~\ref{fig:Architecture}, we need first to populate a 5-way tensor $T\in \mathbb{R}^{N_{id}\times D_y\times D_p\times D_r\times D_f}$ with $N$ samples $x_n\in\mathbb{R}^{D_f}$ of the training set containing $N_{id}$ number of subjects (identities) where $x_n$ refers to a $D_{f}$-dimensional vector representing features (in our case 3D landmark points) extracted from the face image. These $x_n$ are generated by intrinsically rotating 3D models of 300 subjects from facescape dataset~\cite{yang2020facescape} for the selected poses. Therefore, each $x_n$ is labelled according to its identity and 3 values defining its corresponding rotation angles (i.e., yaw, pitch and roll) in degrees. 
These rotation angles can be discretized into $D_y$, $D_p$ and $D_r$ bins respectively. To populate the tensor, we normalize the extracted landmarks ($L$) by first computing their centroid ($\mathbf{C}$), then calculating the scaling factor ($s$) as shown in Eq.~\ref{eq:landmark_normalization}, and dividing all landmarks by this factor. This ensures that all face images share the same scale and translation.

\begin{equation}
  \begin{array}{l}
    L = \{ (x_i, y_i, z_i) \}_{i=1}^n \\[10pt]
    \mathbf{C} = \left( \dfrac{1}{n} \sum_{i=1}^{n} x_i, \dfrac{1}{n} \sum_{i=1}^{n} y_i, \dfrac{1}{n} \sum_{i=1}^{n} z_i \right) \\[15pt]
    \| \mathbf{L}_i - \mathbf{C} \|^2 = (x_i - C_x)^2 + (y_i - C_y)^2 + (z_i - C_z)^2 \\[10pt]
    s = \sqrt{\dfrac{1}{n} \sum_{i=1}^{n} \| \mathbf{L}_i - \mathbf{C} \|^2} \\[14pt]
    \hat{L}_i = \left( \dfrac{x_i- C_x}{s}, \dfrac{y_i- C_y}{s}, \dfrac{z_i- C_z}{s} \right)
  \end{array}
  \label{eq:landmark_normalization}
\end{equation}

\subsection{Tensor decomposition and optimizing sinusoidal parameters}
\label{subsec:tensor-decomposition}
Once the tensor $T$ is populated, we can decompose it using Eq.~\ref{eq:tucker_modeproduct} as:
\begin{equation}
T = G\times_1 A^{(id)}\times_2 A^{(y)}\times_3 A^{(p)}\times_4 A^{(r)}\times_5 A^{(f)}\\[8pt]
\label{eq:tensor_decomposition}
\end{equation}

Each factor matrix $A^{(*)}$ spans a subspace corresponding to the given factor. Consequently, its rows, can be interpreted as vectors that represent the data's behavior across each parameter within that factor's subspace. Its columns can be viewed as component (or factor vector) contributing to the overall structure or pattern of the data along its corresponding mode (dimension) of the tensor. For example, the rows of the matrix $A^{(id)}$ denoted as $a^{(*)}$ captures the distinctive features that characterize the shape of the object $x_n\in\mathbb{R}^{D_f}$. Meanwhile, each row in the matrices $A^{(y)}, A^{(p)}, A^{(r)}$ contains coefficients that determine the object's rotation by a specific angle around each axis. Since the n-mode product of $G$ by $A^{(f)}$ can be interpreted as a mapping of the shaped and rotated object into feature space, they can be combined into a auxiliary variable $\mathbf{W} = G\times_1A^{(f)}$~\cite{vasilescu2002multilinear}.

After decomposing the tensor $T$, a sample $x$ can be reconstructed, denoted as $\bar{x}$, as:
\begin{equation}
x \simeq \bar{x} = \mathbf{W}\times a^{(id)}\times a^{(y)}\times a^{(p)}\times a^{(r)}\\[10pt]
\label{eq:sample_reconstruction}
\end{equation}
where \{$a^{(id)}, a^{(y)}, a^{(p)}, a^{(r)}$\} are row vectors from factor matrices {$A^{(id)}, A^{(y)}, A^{(p)}, A^{(r)}$}. Therefore, given a test sample $x\in\mathbb{R}^{D_f}$, rotation angles can be estimated theoretically~\cite{tenenbaum1996separating} by minimizing the reconstruction error as:

\begin{equation}
\operatorname*{argmin}_{a^{(id)}, a^{(y)}, a^{(p)}, a^{(r)}} \lVert x-\mathbf{W}\times a^{(id)}\times a^{(y)}\times a^{(p)}\times a^{(r)} \rVert\\[10pt]
\label{eq:sample_reconstruction}
\end{equation}

This is a minimization problem where we need to simultaneously optimize four vectors to accurately estimate yaw, pitch, and roll for pose estimation. While various methods~\cite{bakry2014untangling} exist to solve this minimization, achieving precise results is challenging, and the obtained solutions often fail to preserve the manifold structure of the different subspaces. Therefore, the minimization results must be constrained to align with the manifold structure inherently defined by the training examples.
Using the Eq.~\ref{eq:tensor_decomposition} to decompose a tensor, we obtain three different matrices: $A^{(y)}$, $A^{(p)}$, and $A^{(r)}$, spanning the sub-spaces corresponding to rotation about yaw, pitch, and roll respectively. According to~\cite{derkach2019tensor}, if we plot the values of the columns of these matrices, they approximately form a spiral curve that can be well approximated through cosine functions. Thus, they reveal that the coefficients of the rotation subspace follow a uni-modal manifold structure. Building on this observation, explicit constraints can be applied to the rotation coefficients, enabling their joint estimation directly on the underlying rotation manifold. Therefore, Eq.~\ref{eq:sample_reconstruction} can be re-written as follows:
\begin{equation}
\begin{array}{l}
\underset{\omega^{(id)}, \omega^{(y)}, \omega^{(p)}, \omega^{(r)}}{\operatorname{argmin}} \lVert x - \hat{x} \rVert \\[10pt]
\hat{x} = \mathbf{W} \times a^{(id)} \times f^{(y)}(\omega^{(y)}) \times f^{(p)}(\omega^{(p)}) \times f^{(r)}(\omega^{(r)}) \\[10pt]
\end{array}
\label{eq:constrained_reconstruction}
\end{equation}
where $f^{(*)}:\mathbb{R}\rightarrow\mathbb{R}^{D_*}$ are trigonometric functions taking as input an angle $\omega^{(*)}$ and giving as output a vector of coefficients $a^{(*)}$. $D_*$ represents number of dimensions given a subspace, which corresponds to the number of columns in each of the factor matrices. Both the parameters $\omega^{(*)}$ and $a^{(*)}$ are optimized variables. Therefore, $f^{(*)}$ will be represented as a vector of real functions parameterized by cosine, as follows:
\begin{equation}
  \begin{array}{l}
    f^{(*)}(\omega^{(*)}) = (f_1(\omega^{(*)}), f_2(\omega^{(*)}),...,f_{D_*}(\omega^{(*)})) \\[10pt]
    f_j(\omega^{(*)}) = \alpha_j^{(*)}cos(\beta_j^{(*)}\omega^{(*)}+\gamma_j^{(*)})+\varphi_j^{(*)}\\[10pt]
    1\leq j<D_*
    
  \end{array}
  \label{eq:trignometric_function}
\end{equation}
where $\alpha_j^{(*)},\beta_j^{(*)},\gamma_j^{(*)}$ and $\varphi_j^{(*)}$ are sinusoidal parameters that control the behavior of a trigonometric wave. It is important to note that for each rotation subspace, a unique set of parameters will be associated with each dimension of the subspace. The values of these sinusoidal parameters are initialized using Fourier transform and obtained by optimizing the following minimization:
\begin{equation}
\underset{\alpha_j^{(*)},\beta_j^{(*)},\gamma_j^{(*)},\varphi_j^{(*)}}{\operatorname{argmin}} \lVert a_{ij}^{(*)} - f_j(\omega_{i}^{(*)}) \rVert \\[10pt]
\label{eq:constrained_reconstruction}
\end{equation}
where $a_{ij}^{(*)}$ are the elements of matrices $U^{(y)}, U^{(p)}, U^{(r)}$, which are derived from the decomposition of the trained tensor. The term $\omega_i^{(*)}$ represents the angle value corresponding to the \textit{i}-th bin of the discretized rotation angles used in the tensor's construction. 

\begin{table*}[bp!]
    \centering
    \caption{Comparison with SOTA methods in terms of mean absolute error of Euler angles and vectors on the AFLW2000 and BIWI datasets}
    \label{tab:BIWI_AFLW_errors}
    \renewcommand{\arraystretch}{1.2}  
    \setlength{\tabcolsep}{1.4pt}  
    \resizebox{1.0\textwidth}{!}{%
    \begin{tabular}{l|ccc>{\columncolor{gray!20}}c|ccc>{\columncolor{gray!20}}c|ccc>{\columncolor{gray!20}}c|ccc>{\columncolor{gray!20}}c} 
        \hline
        \multirow{3}{*}{Method} & \multicolumn{8}{c|}{\textbf{Euler angles errors}} & \multicolumn{8}{c}{\textbf{Vector errors}} \\  
        \cline{2-17} 
         & \multicolumn{4}{c|}{AFLW2000} & \multicolumn{4}{c|}{BIWI} & \multicolumn{4}{c|}{AFLW2000} & \multicolumn{4}{c}{BIWI} \\ 
        \cline{2-17}
        & Yaw & Pitch & Roll & MAE & Yaw & Pitch & Roll & MAE & Left & Down & Front & MAEV &  Left & Down & Front & MAEV \\ 
        \hline
        3DDFA~\cite{zhu2016face} & 4.71 & 27.08 & 28.43 & 20.07 & 5.50 & 41.90 & 13.22 & 20.20 & 30.57 & 39.05 & 18.52 & 29.38 & 23.31 & 45.00 & 35.12 & 34.47 \\
        
        Dlib~\cite{kazemi2014one} & 8.50 & 11.25 & 22.83 & 14.19 & 11.86 & 13.00 & 19.56 & 14.80 & 26.56 & 28.51 & 14.31 & 23.12 & 24.84 & 21.70 & 14.30 & 20.28 \\
        
        HopeNet~\cite{ruiz2018fine} & 5.31 & 7.12 & 6.13 & 6.19 & 6.00 & 5.88 & 3.72 & 5.20 & 7.07 & 5.98 & 7.50 & 6.85 & 7.65 & 6.73 & 8.68 & 7.69\\
        
        FSA-Net~\cite{yang2019fsa} & 4.96 & 6.34 & 4.77 & 5.36 & 4.56 & 5.21 & 4.56 & 4.28 & 6.75 & 6.21 & 7.34 & 6.77 & 6.03 & 5.96 & 7.22 & 6.40 \\
        
        QuatNet~\cite{hsu2018quatnet} & 3.97 & 5.62 & 3.92 & 4.50 & 2.94 & 5.49 & 4.01 & 4.15 & - & - & - & - & - & - & - & - \\
        
        HPE~\cite{huang2020improving} & 4.80 & 6.18 & 4.87 & 5.28 & 3.12 & 5.18 & 4.57 & 4.29 & - & - & - & - & - & - & - & - \\
        
        TriNet~\cite{cao2021vector} & 4.36 & 5.81 & 4.51 & 4.89 & \textbf{3.11} & \textbf{5.09} & 5.20 & 4.47 & 6.16 & 5.95 & 6.82 & 6.31 & 6.58 & \textbf{5.80} & 7.55 & 6.64 \\
        \hline                
        TokenHPE~\cite{zhang2023tokenhpe} & \textbf{2.68} & 3.41 & \textbf{1.59} & \textbf{2.56} & 4.06 & 5.33 & \textbf{2.41} & 3.93 & \textbf{3.38} & 3.90 & \textbf{4.63} & \textbf{3.97} & 5.21 & \textbf{5.71} & 7.06 & 6.00 \\
        
        6DRepNet~\cite{hempel2024toward} & 2.79 & \textbf{3.39} & 1.65 & 2.61 & 3.43 & 5.22 & 2.61 & \textbf{3.75} & 3.47 & \textbf{3.87} & 4.71 & 4.02 & \textbf{4.77} & 5.72 & \textbf{6.48} & \textbf{5.65} \\
        
        NLML-HPE (ours) & 3.06 & 4.23 & 1.96 & 3.08 & 3.58 & 5.29 & 2.67 & 3.85 & 4.02 & 4.77 & 5.53 & 4.78 & 5.34 & 6.03 & 6.63 & 6.00 \\
        \hline
    \end{tabular}    
    }
\end{table*}

\subsection{Training of the encoder and MLP heads}
\label{subsec:train}
Solving the optimization problem in Eq.~\ref{eq:constrained_reconstruction} is highly time-consuming and impractical for real-time pose estimation. Instead, we train a lightweight encoder that maps flattened facial landmark points (1404 features) into 9 target variables. The encoder architecture is a fully connected feedforward neural network composed of six linear layers with progressively decreasing sizes: 1024, 512, 256, 128, 64, and 9. ReLU activation functions are applied after each intermediate layer, except for the penultimate one, which uses a Tanh activation. The 9 outputs are grouped into three sets of three, representing the predicted factor vectors $\hat{a}^{(y)}, \hat{a}^{(p)}, \hat{a}^{(r)}$, corresponding to yaw, pitch, and roll, respectively. The factor vectors within the factor matrices $A^{(y)}, A^{(p)}, A^{(r)}$ serve as the ground truth for this mapping. We also train three MLP heads, each of which takes one of these factor vectors as input and predicts the corresponding Euler angle. All three MLP heads share the same architecture. To generate fine-grained factor vectors (i.e., vectors with very small angle intervals), we use the pre-trained sinusoidal parameters $(\alpha_j^{(*)},\beta_j^{(*)},\gamma_j^{(*)},\varphi_j^{(*)})$. These fine-grained factor vectors serve as ground truth data for training the MLP heads.

During testing, the encoder and MLP heads are concatenated sequentially into a single model, enabling real-time head pose prediction for a given face image. A key advantage of this approach over existing methods is its ability to learn the mapping from facial features to pose latents with limited training data. This is made possible by the accurate approximation of rotation subspaces using tensor decomposition. Consequently, even with a small dataset, a compact yet effective encoder can be trained for this mapping. Furthermore, unlike previous methods, our model delivers real-time head pose estimation. We evaluated its computational efficiency against SOTA approaches on a Core i7-13700K CPU. As shown in Table~\ref{tab:facescape_errors} our method requires only $1.38\times10^{-3}$ seconds per frame to predict the pose, whereas TokenHPE~\cite{zhang2023tokenhpe} and 6DRepNet~\cite{hempel2024toward} take $16.88\times10^{-3}$ and $17.59\times10^{-3}$ s/frame respectively.

\section{Experiments}
\label{sec:experiments}
\subsection{Implementation Details}
\label{subsec:ImplementationDetails}
We begin our experiments by creating a pose-consistent dataset, where rotation angles are discretized in 10-degree intervals. Specifically, yaw angles range from $-50^\circ$ to $+50^\circ$, pitch angles from $-40^\circ$ to $+40^\circ$, and roll angles from $-30^\circ$ to $+30^\circ$. As discussed in Sec.~\ref{subsec:pose-consistent}, we use the combination of these angles to generate our dataset by intrinsically rotating
3D models of 300 subjects from Facescape dataset~\cite{yang2020facescape}. We have defined these bounds due to the limit of mediapipe face mesh to detect facial landmarks at extreme poses. This can be compensated by multiplying the given rotation angle with the facial landmarks of the frontal face (i.e., when yaw, pitch, and roll are all $0^\circ$). However, our experiments shows that this multiplication is very complicated and may not result in the same landmark pose that mediapipe extracts. A better alternative might be to use transformers~\cite{vaswani2017attention} to extract features and fill the tensor with fewer limits on angles bounds. 

After performing the decomposition, we utilize the leading three components (i.e., dimensions) of the factor matrices, denoted as $a_j^{(y)}, a_j^{(p)}, a_j^{(r)}$ for $1\leq j<D_*$ to train the encoder. These dimensions collectively capture approximately 95\% of the total energy (majority of the data's variance), while the remaining dimensions primarily represent noise.

To reduce the computational cost of tensor decomposition, we set the interval of rotation angles in factor matrices to 10 degress. However, this coarse interval may not provide sufficient representation to accurately map the encoder's latent poses to their corresponding Euler angles in the MLP heads. To address this without altering the tensor decomposition, we fit the cosine function in Eq.~\ref{eq:trignometric_function} to each factor vector, optimizing the sinusoidal parameters ($\alpha_j^{(*)},\beta_j^{(*)},\gamma_j^{(*)}, \varphi_j^{(*)}$). With these parameters, we generate new factor matrices with finer angle intervals (e.g., 0.01 degrees). These refined factor matrices are then used to efficiently train the MLP heads.

\subsection{Datasets and Evaluation}
\label{subsec:Datasets}

We conduct our experiments by training on 70\% of our generated dataset, which was synthesized using the FaceScape dataset~\cite{yang2020facescape}. For testing, we use two popular public benchmark datasets: AFLW2000~\cite{zhu2015high} and BIWI~\cite{fanelli2013random}, along with the remaining 30\% of our generated dataset.

\textbf{The FaceScape dataset}~\cite{yang2020facescape} is a large-scale collection of 3D facial models designed for research in 3D face reconstruction and related fields. It comprises 18,760 textured 3D face models captured from 938 subjects, each performing 20 specific expressions

\textbf{The AFLW2000 dataset}~\cite{zhu2015high} dataset consists of the first 2,000 images from the AFLW dataset [18]. It features a diverse range of facial appearances and background settings
\textbf{The BIWI dataset}~\cite{fanelli2013random} contains 15,678 images of 20 individuals (6 females and 14 males), with 4 individuals appearing twice. The head pose range spans approximately $\pm 75^\circ$ in yaw and $\pm 60^\circ$ in pitch.

In order to compare to the state-of-the-art, we adopt Mean Absolute Errors (MAE) of the Euler angles and Mean Absolute Error of the rotation matrix vector (MAEV) as evaluation metrics with the same test setting as mentioned in TokenHPE~\cite{zhang2023tokenhpe} and 6DRepNet~\cite{hempel2024toward}.

\subsection{Experiment Results}
\label{subsec:ExperimentResults}
We evaluate NLML-HPE against state-of-the-art methods on public benchmarks under identical experimental conditions, using MAE and MAEV as metrics. For earlier models, we report MAE and, when available, MAEV from their original papers. MAEV, proposed by Cao et al.\cite{cao2021vector}, addresses the discontinuity of Euler angles. Table\ref{tab:BIWI_AFLW_errors} shows results on the AFLW2000 and BIWI datasets. While our method was trained on our custom dataset, all other methods were trained on the 300W-LP dataset. Our method demonstrates strong generalization across both MAE and MAEV metrics, consistently achieving competitive performance. On the AFLW2000 dataset, NLML-HPE achieves an average MAE of 3.08, outperforming Dlib (14.19), 3DDFA (20.07), HopeNet (6.19), and FSA-Net (5.36), and showing particularly accurate roll estimation ($1.96^\circ$) better than FSA-Net ($4.77^\circ$) and close to TokenHPE ($1.59^\circ$). Although NLML-HPE slightly trails TokenHPE (2.56) and 6DRepNet (2.61), it compensates with consistent performance across error types. In vector-based evaluation, NLML-HPE achieves a MAEV of 4.78, outperforming TriNet (6.31), HopeNet (6.85), and FSA-Net (6.77), and remaining competitive with TokenHPE (3.97) and 6DRepNet (4.02), highlighting its geometrically consistent and robust 3D pose estimation.

On the BIWI dataset, NLML-HPE maintains strong performance with MAE of 3.85, falling just short of 6DRepNet (3.75) while outperforming TriNet (4.47) by approximately 13.8\% and showing improvements over most earlier methods. Compared to TokenHPE (3.93), NLML-HPE achieves better pitch estimation (5.29 vs. 5.33) and competitive yaw accuracy (3.58 vs. 4.06), indicating balanced prediction across angles. In vector-based evaluation, NLML-HPE attains a MAEV of 6.00, surpassing TriNet (6.64) and FSA-Net (6.40), and remaining close to 6DRepNet (5.65). These results highlight NLML-HPE's robust and consistent pose estimation, with lower directional variance and competitive accuracy despite using less training data and a less complex architecture.

\begin{table}[h!tbp] 
    \centering
    \caption{Comparison with SOTA methods in mean absolute errors of Euler angles, vector errors, and computation time per frame on our validation set. Y: Yaw, P: Pitch, R: Roll, L: Left, D: Down, F: Front, TPF: time per frame.}
    \label{tab:facescape_errors}
    \renewcommand{\arraystretch}{1.2}  
    \setlength{\tabcolsep}{0.5pt}  
    \resizebox{0.48\textwidth}{!}{%
    \begin{tabular}{l|c|c|c>{\columncolor{gray!20}}c|c|c|c>{\columncolor{gray!20}}c|c}
        \hline
        \multirow{2}{*}{Method} & \multicolumn{4}{c|}{Euler angles errors}                       & \multicolumn{4}{c|}{Vector errors}       & \multirow{2}{*}{TPF (ms)}                              \\  
        \cline{2-9}
        & Y & P & R & MAE & L & D & F & MAEV & \\ 
        \hline
                
        TokenHPE~\cite{zhang2023tokenhpe} & 9.0 & 6.1 & 6.3 & 7.1 & 11.0 & 8.4 & 11.2 & 10.2 & 16.88
        \\
        
        6DRepNet~\cite{hempel2024toward} & 28.0 & 13.6 & 6.3 & 16.0 & 30.2 & 13.8 & 32.3 & 25.4 & 17.59\\
        
        NLML-HPE (ours) & \textbf{3.4} & \textbf{3.5} & \textbf{2.4} & \textbf{3.1} & \textbf{4.4} & \textbf{4.5} & \textbf{5.0} & \textbf{4.6} & \textbf{1.16}\\
        \hline
    \end{tabular}    
    }
\end{table}

\begin{figure*}[h!tbp]
 \centering 
 \includegraphics[trim={0cm 5cm 0 5.5cm},clip,width=175mm,scale=0.5]{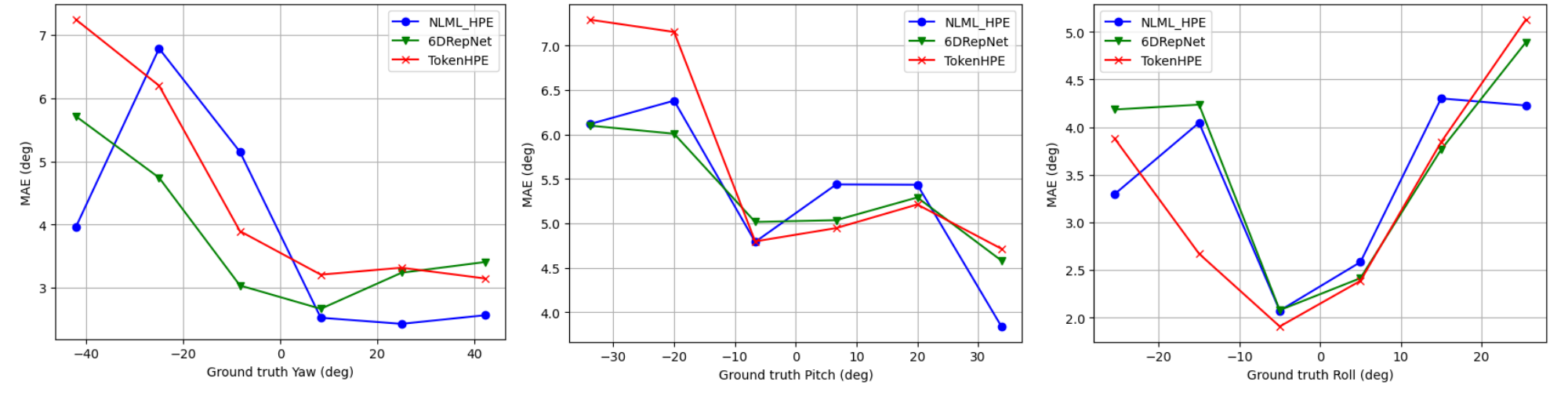}
 \caption{Error analysis for angle intervals on the BIWI dataset}
 \label{fig:ErrorBIWI}
\end{figure*}

\begin{figure*}[h!tbp]
 \centering 
 \includegraphics[trim={0cm 5cm 0 5.5cm},clip,width=175mm,scale=0.5]{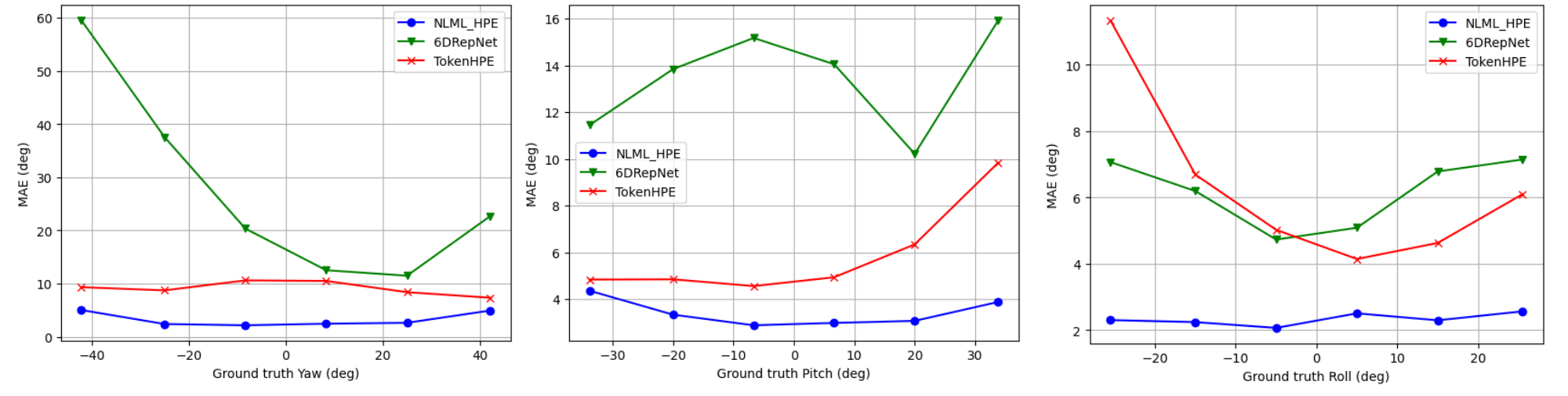}
 \caption{Error analysis for angle intervals on our validation set}
 \label{fig:ErrorFacescape}
\end{figure*}

Comparing our model with SOTA methods over our validation set, as shown in Table~\ref{tab:facescape_errors}, NLML-HPE outperforms these methods in both Euler angle and vector error metrics, particularly in its ability to generalize to unseen data. NLML-HPE achieves a MAE of 1.8, substantially lower than TokenHPE's 16.3 and 6DRepNet's 29.0, highlighting its superior accuracy in predicting Euler angles. This performance is further reflected in the MAEV, where NLML-HPE achieves a value of 3.0, compared to TokenHPE's 25.7 and 6DRepNet's 37.9, showing its robust ability to accurately estimate 3D pose. 

While TokenHPE and 6DRepNet perform well on commonly used datasets such as AFLW2000 and BIWI, which often share patterns between training and validation data, they struggle to generalize effectively to diverse data, a challenge that NLML-HPE effectively overcomes, making it better suited for real-world scenarios. 


\subsection{Error Analysis}
\label{subsec:ErrorAnalysis}
We conduct the error analysis of our proposed NLML-HPE method and those of SOTA, TokenHPE and 6dRepNet on BIWI dataset and our validation set. These results are illustrated in Fig.~\ref{fig:ErrorBIWI} and Fig.~\ref{fig:ErrorFacescape} respectively. In these plots, the yaw angle range of $[-50^\circ, +50^\circ]$ is uniformly divided into intervals of $16.67^\circ$, the pitch range of $[-40^\circ, +40^\circ]$ into intervals of $13.33^\circ$ and the roll range of $[-30^\circ, +30^\circ]$ into intervals of $10^\circ$. 

Looking at these plots, the first noticeable issue is the generalization problem of 6dRepNet and TokenHPE when applied to our generated dataset, as evidenced by the high prediction errors in Fig.~\ref{fig:ErrorFacescape}. Although TokenHPE reports closest MAE to ours for yaw and pitch poses with a smaller deviation from the center, the error increases significantly for other angles, particularly at extreme poses. 
In contrast, our method, trained on a limited custom generated set, achieves nearly consistent MAE across all Euler angles, as shown in Fig.~\ref{fig:ErrorBIWI}, even outperforming the SOTA methods in extreme poses with lower MAE.
This suggests that our NLML-HPE method is better suited for real-world scenarios, as it combines a lightweight architecture with lower prediction error on unseen data.

\section{limitation}
\label{sec:limitations}
As discussed in Section~\ref{subsec:ImplementationDetails}, our experiments were conducted with yaw in the range of $[-50^\circ, +50^\circ]$, pitch in $[-40^\circ, +40^\circ]$, and roll in $[-30^\circ, +30^\circ]$. We restricted our predictions to these ranges due to the limitations of our feature extractor, which struggles to accurately capture facial landmarks at poses with large deviation from the frontal pose. However, this limitation can be overcome with a more accurate feature extractor capable of extracting features in extreme poses.

\section{Conclusion}
\label{sec:conclusion}

In this paper, we propose a novel head-pose estimation algorithm based on non-linear manifold learning. This approach maps the problem of pose estimation in 3D space to the task of learning the underlying structure of separate manifolds. We achieve this by applying multi-linear decomposition to separate the factors of pose variation into distinct subspaces, each corresponding to the effects of yaw, pitch, and roll. We build upon the observation that these subspaces define a continuous curve, which can be modeled using a trigonometric function. By minimizing this function, we obtain solutions that are always consistent with variations in the rotation parameters. Later, we train a deep encoder along with three MLP heads, using these solutions as ground truth to perform real-time pose estimation. Experimental results show that NLML-HPE achieves near state-of-the-art performance on common datasets, despite being trained on a limited, custom-generated dataset. In future work, we plan to extend our method using more accurate feature extractors capable of capturing facial features across all rotation angles.


\section*{Acknowledgment}
This work has been funded by the European Union (GA 101119800 - EMERALD). 

{\small
\bibliographystyle{ieee}
\bibliography{egbib}
}

\end{document}